# Variation in prediction accuracy due to randomness in data division and fair evaluation using interval estimation: By analyzing F1 scores of 33600 machine learning models

**Isao goto[1]**

[1] School of Food Industrial Sciences, Department of Food Resource Development, Miyagi University, 2-2-1 Hatatate, Taihaku-Ku, Sendai, Miyagi; gotoui@myu.ac.jp

**Abstract:** This paper attempts to answer a "simple question" in building predictive models using machine learning algorithms. Although diagnostic and predictive models for various diseases have been proposed using data from large cohort studies and machine learning algorithms, challenges remain in their generalizability. Several causes for this challenge have been pointed out, and partitioning of the dataset with randomness is considered to be one of them. In this study, we constructed 33,600 diabetes diagnosis models with "initial state"-dependent randomness using autoML (automatic machine learning framework) and open diabetes data, and evaluated their prediction accuracy. The results showed that the prediction accuracy had an initial state-dependent distribution. Since this distribution could follow a normal distribution, we estimated the expected interval of prediction accuracy using statistical interval estimation in order to fairly compare the accuracy of the prediction models.

## 1. Introduction

In recent years, research has been conducted on the use of machine learning algorithms (MLAs), including deep learning techniques, to diagnose various diseases and predict future risks [1-3]. These studies have been accelerated by 1) the increasing sophistication of information and communication technology, 2) large-scale data obtained through longitudinal studies, etc., and 3) the opening of program codes for building predictive models using machine learning. In particular, these studies have become even more active in recent years with the advent of automated machine learning framework [4-6].

As an example, published studies have applied MLA to data from the UK Biobank large longitudinal cohort study to develop models to diagnose and predict disease onset in advance [4, 7]. Such studies have been conducted previously, and in 1988, J. W. Smith et al. applied neural networks to data collected by the National Institute of Diabetes and Digestive and Kidney Diseases from a population of Pima Indians near Phoenix, Arizona, to predict the onset of diabetes [8-11]. This dataset, called the PID dataset, is still the primary dataset used to evaluate MLA in recent years, and in 2014, a method was proposed to combine multiple prediction models to predict onset of the disease, showing a very high prediction accuracy of 0.97 [12-17].

As mentioned above, a great deal of research has been published in recent years on predictive models of disease using machine learning. However, there are issues such as inadequate reporting of prediction models and lack of external validation [18]. The low generalizability of prediction models due to these issues is also a general challenge for machine learning-based prediction models, which can be attributed to the lack of training/evaluation using data sets of specific populations and the lack of fairness in the evaluation of prediction models. In general, the usual procedure when building a prediction model is to randomly split data into training and test data based on initial states in order to evaluate the robustness of the model, and then evaluate the performance of the prediction model after learning with the unlearned test data. Therefore, if the potential properties of the training and test data differ, prediction becomes more difficult, and this division has a significant impact on model evaluation and generalizability. The impact of this initial states on generalizability is (too) obvious and has not been considered in most studies. Similarly, the effect of hyperparameter tuning (HPT) of initial states and MLAs has not been well reported.

The purposes of this study are 1) to clarify the effect of the initial state (not the initial value) on metrics, which is a "simple question" when building prediction models using MLAs, and 2) to propose a method for fair evaluation of prediction models.

## 2. Materials and Methods

Dataset: The PID dataset was used in this study. Although larger and more complex diabetes datasets now exist, this dataset remains the benchmark for diabetes classification studies. In this study, this dataset was downloaded from kaggle, one of the online data repository sites. The dataset is from a population of Pima Indian women near Phoenix, Arizona, USA, which has a high incidence of diabetes, with 768 data records, including 268 individuals (about 35%) with diabetes [9-11, 19]. The risk factors for diabetes included in this dataset and the Statistics for each characteristic are shown in Table 1. Outcome is binary values of 0 (not diabetes mellitus) and 1 (diabetes mellitus), and all other risk factors are continuous values. Since the purpose of the present study was not to improve prediction accuracy, no missing value processing was performed.

Table 1. Overview of Pima Indian diabetes dataset.

| Attributes | Property description | Mean | SD | Min | Median | Max |
|---|---|---|---|---|---|---|
| Pregnancies | Number of times pregnant | 3.85 | 3.37 | 0 | 3 | 17 |
| Glucose | Plasma glucose concentration at 2 h in an oral glucose tolerance test | 120.89 | 31.97 | 0 | 117 | 199 |
| Blood Pressure | Diastolic blood pressure (mmHg) | 69.11 | 19.36 | 0 | 72 | 122 |
| Skin Thickness | Triceps skin fold thickness (mm) | 20.54 | 15.95 | 0 | 23 | 99 |
| Insulin | 2-h serum insulin (iU/mL) | 79.80 | 115.24 | 0 | 31 | 846 |
| BMI | Body mass index (weight in Kg / (height in m)$^2$) | 32.0 | 7.9 | 0.0 | 32.0 | 67.1 |
| Diabetes Pedigree Function | Diabetes pedigree function | 0.47 | 0.33 | 0.08 | 0.37 | 2.42 |
| Age | Age (years) | 33.24 | 11.76 | 21 | 29 | 81 |
| Outcome | diabetic population marker | 0.35 | 0.48 | 0 | 0 | 1 |

AutoML framework: Python 3.10 and the automatic machine learning framework PyCaret 3.2.0 [20] were used to build the 33600 prediction models used in this study. PyCaret has over 20 MLAs that can be used for classification problems. In this study, a total of 12 MLAs were used, grouped into three categories, as shown in Table 2. PyCaret also has various functions to streamline the process of building predictive models, and in this study, the setup() function was used for data preprocessing, the create_model() function for model building, the tune_model() function for tuning hyperparameters, the predict predict_model() function was used for prediction.

·setup() performs preprocessing (data preparation) and environment setting. By specifying the optional parameter session_id for this function, train/test data, which are split randomly, are also reproduced. In this study, Train and Test data are split 7:3. In this paper, session_id (SI) is referred to as "initial states". Other optional parameters, fold_strategy and fix_imbalance, can be used to reduce the type of cross-validation and data imbalance.

·create_model() builds a prediction model by specifying the machine learning algorithm to be used. train data is used to Train and evaluate the model.

·tune_model() automatically tunes the hyperparameters of the prediction model built by create_model(). In this case, default values were used except for the number of tuning times. Train data is used to tune the hyperparameters and is used for evaluation. The function can set tuning iterations and evaluation indicators with optional parameters n_iter and optimize, respectively.

·predict_model() uses Test data to evaluate the performance of the predictive model evaluations built or tuned with create_model() or tune_model().

I also used default values for optional parameters not explicitly stated. Since the objective of this study was not to improve prediction accuracy, no feature engineering was performed. For the same reason, the machine learning algorithm catboost, which performs target encoding, was not used. In this paper, each algorithm is named using the symbols in the table.   In this study, in order to build a prediction model under different conditions, the initial values were set in setup(): session_id was set to a value between 0 and 99, fold_strategy was set to "kfold" or "stratifiedkfold", fix_imbalance was set to "True" or "False". Also, HPT was performed for a total of six conditions, combining two conditions of n_iter (10 and 50) and three conditions of optimize (Accuracy, Recall and Precision). The combination of the above resulted in a final predictive model of 33,600.

Evaluation: The F1 score is the harmonic mean of Recall and Precision, which is the standard metric used when outcomes are unbalanced. Recall is a metric that indicates the proportion of actual positive instances that were correctly predicted, while Precision represents the proportion of predicted positive instances that are actually correct. The F1 score balances these two metrics to provide a comprehensive measure of the model's overall performance. The equations for each metrics are shown below.

$$Recall = \frac{TP}{TP + FN} \tag{1}$$

$$Precision = \frac{TP}{TP + FP} \tag{2}$$

$$F1\ score = \frac{2 * (Precision * Recall)}{Precision + Recall} \tag{3}$$

where in the equation, in the above equation, TP, FN, and FP denote True Positive, False Negative, and False Positive, respectively.

For fair comparison：Even if the condition for any hyperparameters (condition C) are fixed, any number of prediction models can be constructed by changing the initial states. Assuming that the accuracy of the model influenced by the initial states follows a normal distribution, a frequentist statistical analysis in the framework of the sample size design can be used to determine the expected value of the accuracy in condition C.

**Table 2. Machine learning algorithms used in this study.**

| Algorithm name | Symbol | Category | Main parameters to be tuned |
|---|---|---|---|
| Logistic Regression | lr | Linear | C, penalty |
| Ridge Classifier | ridge | Linear | alpha |
| K-Nearest Neighbors | knn | Famous Algorithms | n_neighbors, weights, algorithm |
| Naive Bayes | nb | Famous Algorithms | priors, var_smoothing |
| Support Vector Classifier | svm | Famous Algorithms | C, kernel, gamma, degree |
| Decision Tree Classifier | dt | Tree-based Algorithms | criterion, max_depth, min_samples_split, min_samples_leaf, max_features, ccp_alpha |
| Random Forest Classifier | rf | Tree-based Algorithms (Bootstrap Aggregating) | n_estimators, criterion, max_depth, min_samples_split, min_samples_leaf, max_features |
| Extra Trees Classifier | et | Tree-based Algorithms (Bootstrap Aggregating) | n_estimators, criterion, max_depth, min_samples_split, min_samples_leaf, max_features |
| Gradient Boosting Classifier | gbc | Tree-based Algorithms (Boosting) | n_estimators, learning_rate, max_depth, min_samples_split, min_samples_leaf, max_features |
| AdaBoost Classifier | ada | Tree-based Algorithms (Boosting) | n_estimators, learning_rate, algorithm |
| XGBoost Classifier | xgb | Tree-based Algorithms (Boosting) | n_estimators, learning_rate, max_depth, min_child_weight, gamma, subsample, colsample_bytree, reg_alpha, reg_lambda |
| LightGBM Classifier | lgb | Tree-based Algorithms (Boosting) | n_estimators, learning_rate, num_leaves, max_depth, min_child_samples, subsample, colsample_bytree |

Assuming that the population variance $\sigma^2$ is known, the confidence rate is $1 - \alpha = 0.95$, and the interval width of the confidence interval (i.e., the acceptable range of error) is $\delta$, the required sample size is shown in Eq (4).

$$n \geq \frac{4z_{\frac{\alpha}{2}}^2 \sigma^2}{\delta^2} \quad (4)$$

where $z_{\frac{\alpha}{2}}$ is the critical value of the standard normal distribution at $\frac{\alpha}{2}$. However, since the mother variance is naturally unknown, it is sufficient to find n satisfying Eq. (4) after obtaining $\sigma$ preliminarily, and then to find n satisfying Eq. (5) sequentially using t-distribution while increasing this value sequentially.

$$2t(n-1, \alpha)\frac{c^*\sigma}{\sqrt{n}} \leq \delta \quad (5)$$

c* is shown in Eq. (6).

$$c^* = \frac{\sqrt{2}\Gamma(\frac{\phi+1}{2})}{\sqrt{\phi}\Gamma\frac{\phi}{2}} \quad (6)$$

where $\phi$ and $\Gamma$ are degrees of freedom (n-1) and gamma functions, respectively. The Shapiro-Wilk test was used to test the normality of the F1 score.

## 3. Results

### 3.1. Characteristics of all models built

The frequency and distribution of F1 scores for all constructed 33600 models were examined and their statistics were calculated (Figure 1). The average and standard deviation of the F1 scores for all models constructed was 0.61 ± 0.10.

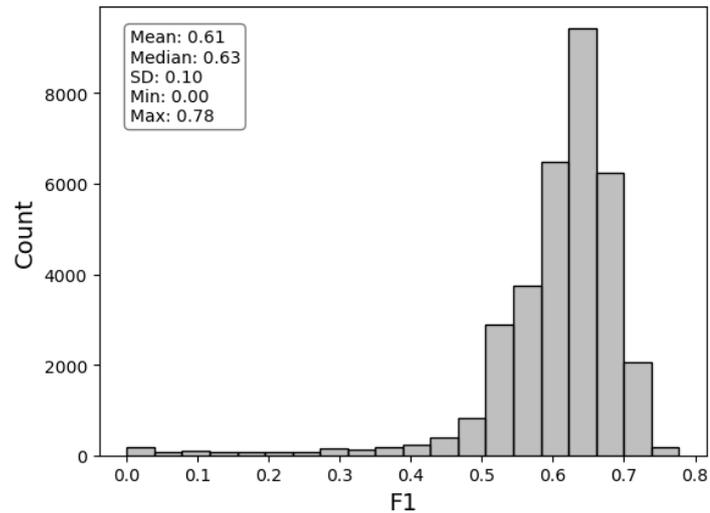

**Figure 1.** Histogram of F1 scores for all models.

### 3.2. The effect of randomness on F1 score

To examine the effect of initial states on F1, subgroups were created for all models constructed, each with a set SI, and the range of F1 values for each subgroup is shown in the boxplot. Thus, each of the 100 subgroups divided by the initial states belongs to 336 prediction models created under different conditions other than the initial states. Figure 2 shows that the median and variability of F1 is different for each subgroup. This result indicates that the initial states used (i.e., Train and Test data split) affect the prediction accuracy.

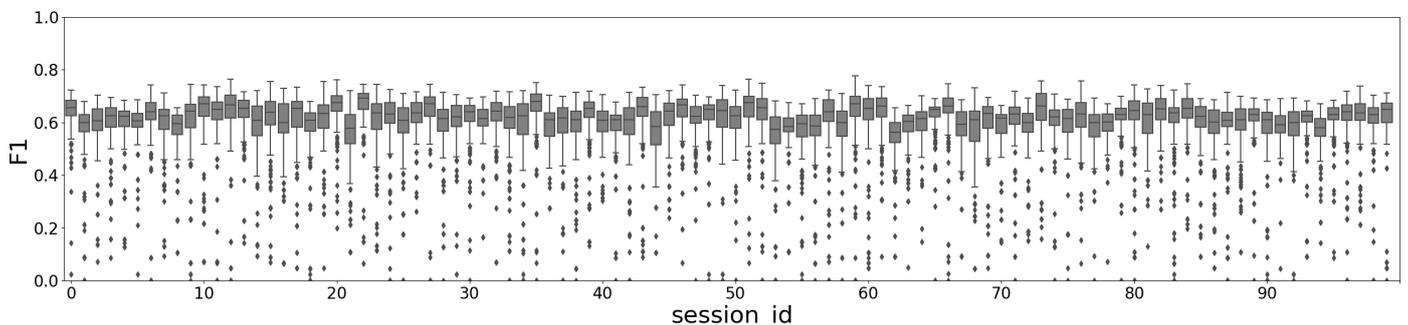

**Figure 2.** Variation in F1 scores for each SI subgroup.

### 3.3. The effect of randomness on each machine learning algorithm

The effect of initial conditions on the 12 MLAs used in this study was examined. Subgroups of data were created with these MLAs, and the maximum, minimum, mean, and range (maximum-minimum) of each F1 and their respective rankings are shown in Table 3. Each MLA subgroup belongs to 2800 prediction models. The maximum prediction accuracy was 0.7640 for Ada, while the minimum was 0.5521. The range of F1 scores within the subgroups was 0.2119. On the other hand, lr's performance was stable in terms of maximum, minimum, mean, and range. Focusing on these two MLAs, each subgroup was further subgrouped by initial values to check the effect of initial values on F1 scores (Figure 3 (a), (b)). Although there are differences in variability between the two algorithms, the impact of initial values on F1 can be confirmed for both MLAs.

**Table 3.** Statistics and ranks of F1 scores for each machine learning algorithm.

| model | max | rank | min | rank | mean | rank | range | rank |
|---|---|---|---|---|---|---|---|---|
| ada | 0.7640 | 1 | 0.5521 | 7 | 0.6523 | 6 | 0.2119 | 10 |
| lr | 0.7485 | 2 | 0.6000 | 1 | 0.6718 | 1 | 0.1485 | 3 |
| ridge | 0.7471 | 3 | 0.5783 | 5 | 0.6716 | 2 | 0.1688 | 8 |
| nb | 0.7345 | 4 | 0.5786 | 4 | 0.6538 | 5 | 0.1559 | 5 |
| rf | 0.7329 | 5 | 0.5823 | 3 | 0.6593 | 3 | 0.1506 | 4 |
| gbc | 0.7294 | 6 | 0.5868 | 2 | 0.6539 | 4 | 0.1426 | 2 |
| et | 0.7241 | 7 | 0.5641 | 6 | 0.6506 | 7 | 0.1600 | 6 |
| xgb | 0.7176 | 8 | 0.5422 | 9 | 0.6285 | 9 | 0.1754 | 9 |
| lgb | 0.7101 | 9 | 0.5443 | 8 | 0.6328 | 8 | 0.1658 | 7 |
| dt | 0.6977 | 10 | 0.4605 | 11 | 0.5746 | 11 | 0.2372 | 11 |
| knn | 0.6557 | 11 | 0.5176 | 10 | 0.5963 | 10 | 0.1381 | 1 |
| svm | 0.5983 | 12 | 0.0000 | 12 | 0.4572 | 12 | 0.5983 | 12 |

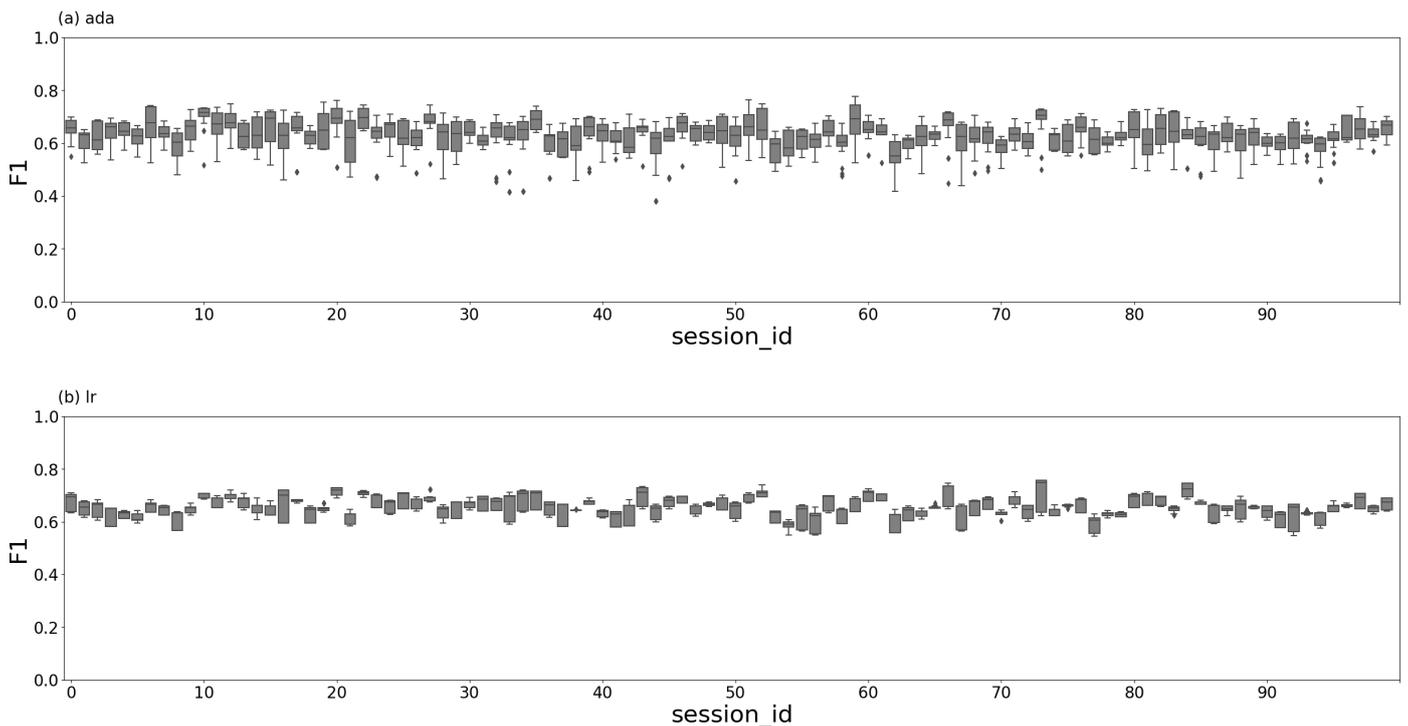

**Figure 3.** Variation in F1 scores for each subgroup of ada (a) and lr (b).

### 3.4. Can hyperparameter tuning reduce the effect of randomness?

Tables 4 and 5 show the mean, standard deviation, maximum, minimum, and range of F1 scores for the Ada and lr subgroups, respectively, when HPT was performed under the six conditions presented in Materials and Methods. The results without HPT are also shown for comparison. For ada, HPT improved prediction accuracy in some conditions, while in others it worsened it. For example, if n_iter is "50" and optimize is "Prec." in ada, the average prediction accuracy drops from 0.6347 to 0.6078. For lr, performance was more stable than for ada, and HPT did not reduce prediction accuracy in most prediction models. These results indicate that regardless of the type of MLA, it is difficult for HPT to prevent the effects of initial conditions on accuracy.

**Table 4.** Statistics of F1 scores for each HPT condition when using ada.

| n_iter | optimize | mean | sd | max | min | range |
|--------|----------|------|------|------|------|-------|
| - | - | 0.6347 | 0.0427 | 0.7640 | 0.5132 | 0.2508 |
| 10 | Accuracy | 0.6411 | 0.0527 | 0.7771 | 0.4918 | 0.2853 |
| 10 | Prec. | 0.6147 | 0.0776 | 0.7771 | 0.3793 | 0.3978 |
| 10 | Recall | 0.6388 | 0.0471 | 0.7558 | 0.4930 | 0.2628 |
| 50 | Accuracy | 0.6403 | 0.0524 | 0.7619 | 0.4677 | 0.2942 |
| 50 | Prec. | 0.6078 | 0.0798 | 0.7619 | 0.3793 | 0.3826 |
| 50 | Recall | 0.6364 | 0.0437 | 0.7447 | 0.4930 | 0.2517 |

**Table 5.** Statistics of F1 scores for each HPT condition when using lr.

| n_iter | optimize | mean | sd | max | min | range |
|--------|----------|------|------|------|------|-------|
| - | - | 0.6504 | 0.0400 | 0.7485 | 0.5455 | 0.2030 |
| 10 | Accuracy | 0.6519 | 0.0394 | 0.7561 | 0.5455 | 0.2106 |
| 10 | Prec. | 0.6509 | 0.0393 | 0.7561 | 0.5455 | 0.2106 |
| 10 | Recall | 0.6729 | 0.0307 | 0.7561 | 0.5783 | 0.1778 |
| 50 | Accuracy | 0.6511 | 0.0392 | 0.7561 | 0.5455 | 0.2106 |
| 50 | Prec. | 0.6502 | 0.0385 | 0.7561 | 0.5455 | 0.2106 |
| 50 | Recall | 0.6719 | 0.0310 | 0.7561 | 0.5714 | 0.1847 |

*3.5. How many models should be created to evaluate a fair model?*

From previous experiments, we expect that the effects of initial conditions on F1 scores are difficult to cancel. Therefore, we believe that it is possible to fairly evaluate the F1 score under a given condition by fixing the conditions other than the initial condition and performing interval estimation from the F1 scores obtained under an arbitrary number of initial conditions. However, since the above requires the assumption that the distribution of F1 scores obtained under the conditions follows a normal distribution, the Shapiro-Wilk test was used. In the present study, the number of predictive models required for interval estimation was calculated under the following conditions.

Conditions:
(Number of prediction models : 3, Machine learning algorithm : "ada", fold_strategy : "kfold", fix_imbalance : "ture", n_iter : "50", optimize : "Accuraccy")

We randomly extracted three F1 scores from the prediction model created under these conditions. The result of the normality test yielded a p-value of 0.38, and since the null hypothesis of the Shapiro-Wilk test, "normally distributed," could not be rejected, we estimated the number of prediction models needed. In accordance with equations (1) to (3), n was calculated by setting $\sigma$ to $\sigma 0$, $\alpha$ to 0.05, and the expected value of the interval width of the confidence interval $\delta$ to 5% of the mean value, 0.0311, which gives 19 as the number of necessary prediction models. From there, sequentially increasing n to 19, the interval estimate of the population mean of the predicted F1 score ranged from 0.6262 to 0.6777. This interval includes the mean of the F1 score (0.6403) for the ada shown in Table 4.

## 4. Discussion

In this study, we used the AutoML framework and PID dataset to investigate the effect of initial conditions on the evaluation metrics F1 score. The influence of the initial state could not be canceled by MLAs or HPT. On the other hand, although in this pilot experiment, I was able to estimate the number of prediction models needed to calculate the expected value of a reliable F1 score by using a general interval estimation technique.

Although many models for diagnosis/prediction of diseases using machine learning have been proposed, not many papers consider initial conditions as in this study. Wu et al. showed the generalizability of the model by evaluating the performance of the predictive model with the mean and standard deviation of the metrics by constructing 10 predictive models with 10 different initial conditions [21]. However, they did not use interval estimation to estimate the expected value of metrics as in this paper.

It is especially important to emphasize that it is not surprising that the initial conditions affect the prediction accuracy. If the characteristics of the train data and the test data, which are divided depending on the initial conditions, are close, the performance of the prediction model is expected to be higher. Therefore, it may be important to evaluate the "similarity" between train and test data. On the other hand, it is difficult to simply discuss the issue of initial conditions alone, since there are known issues in building predictive models, such as over-learning. Attraction of MLAs is the possibility of finding complex combinations among features inherent in the data. Therefore, when validating predictive models, building and evaluating predictive models by a statistically sufficient number of initial conditions is considered very important when building models to diagnose or predict diseases in the future. It is also considered equally important to estimate the sufficient number of initial conditions needed for evaluation.

The purpose of this pilot study is to consider fair evaluation methods to improve the generalizability of predictive models for diagnosis and prediction of diseases using machine learning algorithms. However, at least three limitations of the study include

1) Data set issues:

This study uses a relatively small dataset with a sample size of less than 1000 and less than 10 features. In addition, the study did not perform processes that are important for improving the prediction accuracy of MLAs, such as missing value processing, feature selection, and feature engineering. If the sample size is large enough, the influence of the initial state may be reduced, and the generality of the prediction model using MLAs may likely increase.

(2) Number of initial conditions:

In this present study, the predictive model was built with only 100 initial conditions due to practical problems; since the true population of F1 scores is unknown, it is difficult to argue whether this number was sufficient. If a normal distribution is assumed, the method proposed in this study may be acceptable, but if a normal distribution cannot be assumed, a different approach is needed.

3) MLAs and problems in learning:

Limitations of this study include the fact that only 12 types of machine learning were used and the limited HPT conditions. Similarly, how to deal with the effects of similar initial conditions in regression problems is a future issue.

## 5. Conclusions

This study suggests initial state effects on the performance evaluation of prediction model using MLAs and how to evaluate the model impartially.